%% file: CameraReady.tex
\documentclass[letterpaper]{article} 
\usepackage{aaai23}  
\usepackage{times}  
\usepackage{helvet}  
\usepackage{courier}  
\usepackage[hyphens]{url}  
\usepackage{graphicx} 
\usepackage{natbib}  
\usepackage{caption} 
\usepackage{algorithm}
\usepackage{algorithmic}
\usepackage{amsmath}
\usepackage{color}

\input{macros}

\usepackage{newfloat}
\usepackage{listings}
\usepackage{amssymb}
\usepackage{dsfont}
\usepackage{booktabs}
\usepackage{multirow}
\usepackage{paralist}
\usepackage{pifont}
\usepackage{makecell}
\newcommand{\cmark}{\ding{51}}
\newcommand{\xmark}{\ding{55}}
\usepackage{colortbl}
\usepackage[table]{xcolor}
\usepackage[switch]{lineno}
\DeclareCaptionStyle{ruled}{labelfont=normalfont,labelsep=colon,strut=off} 
\floatstyle{ruled}
\newfloat{listing}{tb}{lst}{}
\floatname{listing}{Listing}
\title{Domain-General Crowd Counting in Unseen Scenarios}
\author{
    Zhipeng Du\textsuperscript{\rm 2}, Jiankang Deng\textsuperscript{\rm3}, Miaojing Shi\textsuperscript{\rm 1,2}\thanks{Corresponding author.\\  \indent\hspace{2mm}Work partially done during Zhipeng Du's internship at Huawei London Research Center.}
}
\affiliations{
    \textsuperscript{\rm 1} College of Electronic and Information Engineering, Tongji University, China \\
    \textsuperscript{\rm 2}King's College London, UK\\ \textsuperscript{\rm 3}Huawei London Research Center, UK\\

    zhipeng.du@kcl.ac.uk, j.deng16@imperial.ac.uk, mshi@tongji.edu.cn
}

\usepackage{bibentry}

\begin{document}

\maketitle

\input{sec0_abstract}

\input{sec1_introduction}

\input{sec2_relatedwork_2}
\input{sec3_method}

\input{sec4_experiments}

\input{sec5_conclusion}

\input{sec6_ack.tex}

\bibliography{aaai23}

\end{document}

%% file: macros.tex
\def \ie {\emph{i.e.}~}
\def \eg {\emph{e.g.}~}


%% file: sec0_abstract.tex
\begin{abstract}
Domain shift across crowd data severely hinders crowd counting models to generalize to unseen scenarios. Although domain adaptive crowd counting approaches close this gap to a certain extent, they are still dependent on the target domain data to adapt (\eg finetune) their models to the specific domain.
In this paper, we aim to train a model based on a single source domain which can generalize well on any unseen domain.  
This falls into the realm of domain generalization that remains unexplored in crowd counting.  
We first introduce a dynamic sub-domain division scheme
which divides the source domain into multiple sub-domains such that we can initiate a meta-learning framework for domain generalization. The sub-domain division is dynamically refined during the meta-learning. 
Next, in order to disentangle domain-invariant information from domain-specific information in image features, we design the domain-invariant and -specific crowd
memory modules to re-encode image features. Two types of losses, i.e. feature reconstruction and orthogonal losses, are devised to enable this disentanglement.  
Extensive experiments on several standard crowd counting benchmarks \ie SHA, SHB, QNRF, and NWPU, show the strong generalizability of our method. {Our code is available at: \url{https://github.com/ZPDu/Domain-general-Crowd-Counting-in-Unseen-Scenarios}} 
\end{abstract}

%% file: sec1_introduction.tex
\section{Introduction}
\label{sec:Introduction}

Crowd counting, which aims to automatically count the number of crowd instances in images/videos, is an essential task in computer vision. Recent years have witnessed numerous progress in crowd counting fueled by the development of deep learning. Most works are learned in a fully-supervised manner~\cite{shi2019revisiting,ma2019iccv,jiang2020cvpr,wang2021iccv,lin2022cvpr} which assumes training and test data are independent and identically distributed.
{In practice, crowd data can be collected from different locations, environments and devices.} These factors cause domain shift in crowd data hence violate the above assumption. 
The conventional fully-supervised models would suffer from severe performance degradation as a result of the distribution mismatch between the training and test data.

To surmount this domain shift in crowd counting, 
many works resort to the domain adaptation~(DA) technique~\cite{wang2019cvpr,sindagi2020eccv,zhao2020active,he2021aaai,gong2022cvpr}. DA focuses on adapting the knowledge learned from the source domain (training) to the target domain (testing) where domain gap exists. Though satisfactory results have been achieved in many DA approaches, they normally rely on the availability of target data to perform model retraining or finetuning.
While in practice, target data are not always available and the target can also change, making DA incompetent.  
\input{Fig/teaser}

In this work, we study an unexplored field, \ie domain generalization (DG), in crowd counting to address the above problem. {As shown in Fig.~\ref{fig:teaser}, }DG aims to learn a model only from the source domain that can generalize to unseen target domains without requiring additional data or model updates. 
Compared to DA, DG is more practical but also more challenging. Its essential idea is to effectively mine domain-invariant knowledge from the source domain data which can be shared and utilized across all domains including unseen ones. 
Meta-learning is a popular approach for solving DG. 
The seminal work, MLDG~\cite{li2018aaai}, 
uses meta-train and meta-test to simulate the domain shift between training and testing phases in reality so as to learn domain-invariant features. Many works follow this optimization strategy to further improve their models' domain generalizability in different tasks~\cite{guo2020cvpr,dai2021cvpr,kim2022cvpr}.

We introduce the first domain-general crowd counting framework. Given a source dataset, based on its intra-dataset variance, we can divide it into multiple sub-domains to simulate them as meta-train/-test sets for meta-learning. Crowd images in different sub-domains contain both domain-invariant (\eg crowd density) and domain-specific (\eg crowd environment) information. 
To learn a generalized model for crowd counting, we need to disentangle the former from the latter in the image feature representation.
We realize this by designing the domain-invariant crowd memory (DICM) and domain-specific crowd memory (DSCM) modules to re-encode each image feature into domain-invariant and -specific features, respectively. It is implemented as a two-branch network, where each consists of a feature projection unit followed by the DICM/DSCM module for feature re-encoding. 
We have one set of memory vectors for DICM that re-encodes all images; and multiple sets for DSCM, each corresponding to one sub-domain re-encodes only images belonging to this domain. To this end, we devise 1) feature reconstruction loss to let re-encoded features be similar to their pre-encoded counterparts; and 2) feature orthogonal loss to let domain-invariant features be dissimilar to their domain-specific counterparts. Finally, the domain-invariant feature is used for crowd density estimation. The network overall optimization follows the meta-learning. We apply a K-means clustering on image features to produce multiple sub-domains. Since image features are gradually optimized during training, we introduce a dynamic sub-domain division scheme to refine our clustering result alongside the update of image features.

In a nutshell, the contribution of this paper is three-fold:

\begin{compactitem}
    \item We introduce the first domain-general crowd counting framework which is trained on a source domain and can generalize well to any unseen target domains. 
    \item We design the domain-invariant and -specific crowd memory modules to disentangle domain-invariant information from domain-specific information in image features. Two types of losses, \ie feature reconstruction and orthogonal losses, are further devised to enable this disentanglement.  
    \item We propose a dynamic sub-domain division scheme which divides the source domain into multiple sub-domains and dynamically refines them for meta-learning optimization. 
\end{compactitem}

We conduct extensive experiments on the SHA, SHB, QNRF and NWPU datasets by training on one and evaluating on the rest. We show our method, without using any target domain data, surpasses state of the art in the same setting; also performs on par with or superior to state of the art which uses target domain data.  

%% file: Fig/teaser.tex
\begin{figure}[t]
\begin{center}

\includegraphics[width=0.9\linewidth]{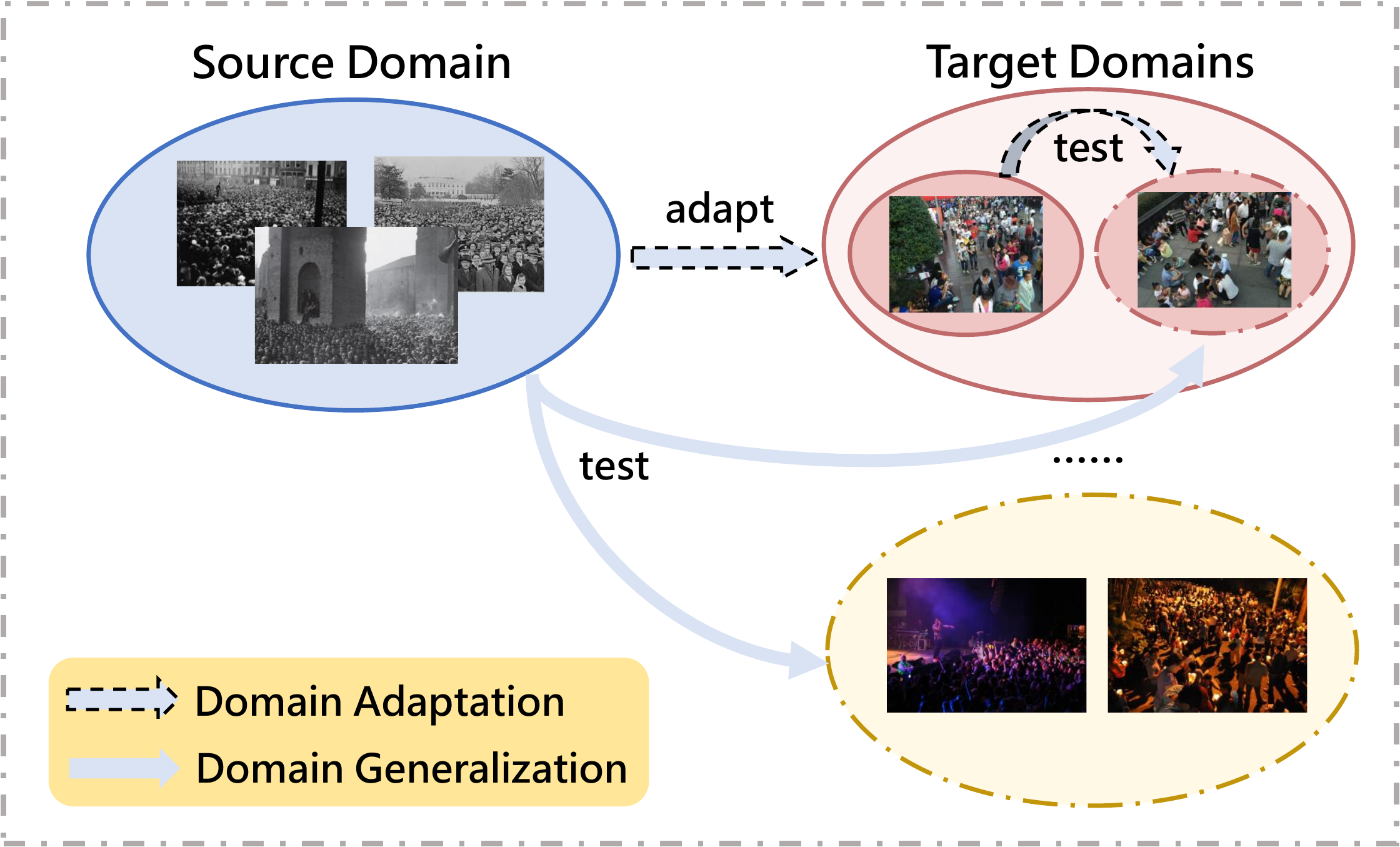}
\end{center}

\caption{Comparison between domain adaptation and domain generalization. The former normally requires data from target domain while the latter does not. } 
%


\label{fig:teaser}

\end{figure}

%% file: sec2_relatedwork_2.tex
\section{Related Work}

\subsection{Crowd Counting } \label{sec:crowdcounting}
Current mainstream approach for crowd counting is to learn a neural network to predict a density map, whose integration indicates the total crowd count in the image. Many works focus on improving model framework~\cite{zhang2018crowd,jiang2020cvpr,liu2021iccv,lin2022cvpr,du2022redesigning}, learning target~\cite{liu2019point,wang2021iccv,song2021iccv,xiong2022eccv}, or introducing novel loss functions~\cite{ma2019iccv,wang2020nips,wan2021cvpr}. These methods bring large improvements to in-distribution test data, but contribute less to out-of-distribution performance. 
{In real world application, there often exists a domain shift between training and test data, which would lead to possible failure of trained models.}

\noindent \textbf{Domain Adaptive Crowd Counting.~} 
To tackle the domain shift problem in crowd counting, many works rely on the domain adaptation (DA) technique~\cite{wang2019cvpr,sindagi2020eccv,he2021aaai,liu2022cvpr,gong2022cvpr}. DA adapts a model learned from a source domain to a target domain, where the target data is normally (at least partially) available, labeled or unlabeled, for model retraining or fine-tuning~\cite{liu2020acmmm,he2021aaai,reddy2021tmm,wu2021mm,liu2022discovering}. For instance, ~\cite{reddy2020wacv,wang2021tnnls,hossain2019bmvc} have developed few-shot learning methods by assuming a few labeled data are available in the target domain for model finetuning. 
The crowd data in the source domain can sometimes be synthetic data~\cite{wang2019cvpr,sindagi2020eccv,zhang2021cvpr,liu2022cvpr,gong2022cvpr}, as they are much easier to obtain and label.  Unlike these works, we propose to generalize a crowd counting model to unseen target domains without relying on target data for model updating. 


\noindent \textbf{Multi-Domain Crowd Counting.} There are also some works studying multi-domain joint learning in crowd counting~\cite{yan2021tcsvt,chen2021variational}. They focus on improving the performance on multiple seen source domains while we focus on improving the performance on any unseen target domain. Their model designs are also fundamentally different from ours. Notice \cite{chen2021variational} also conducts domain division: they train a variational attention module for Gaussian mixture clustering; samples from different datasets are re-grouped for follow-up training. Ours differs from it: we use K-Means clustering and dynamically refine clustering results by taking advantage of updated domain-specific information in image features.

\subsection{Domain Generalization}\label{sec:II-dg}
Domain generalization (DG) methods study the model generalizability on out-of-distribution data. The difference between DA and DG is that DA can access both source and target data while DG only the source data. 
Many DG methods focus on increasing the diversity of the source (training) data~\cite{volpi2018nips,yang2021nips,zhou2021arxiv,wang2021iccv}. For instance,~\cite{volpi2018nips,yang2021nips} augment the training data with adversarially perturbed samples.
Other methods focus on learning domain-invariant knowledge from the source domain~\cite{matsuura2020aaai,lin2021iccv, zhang2022arxiv}. 
Meta-learning has also been popularly used to solve DG. The seminal work, meta-learning domain generalization (MLDG) ~\cite{li2018aaai}, divides the source domain into meta-train and -test sets to simulate the domain shift between training and testing phases. 
Many follow-up works employ this learning manner in different tasks, \eg semantic segmentation~\cite{kim2022cvpr}, face recognition~\cite{guo2020cvpr}, and person re-id~\cite{song2019cvpr,zhao2021cvpr,dai2021cvpr,ni2022cvpr}.  
Besides, memory/prototype networks are
often employed in DG to maintain categorical information across domains~\cite{song2019cvpr,chen2021iccv,kim2022cvpr,dai2021cvpr}. For instance,~\cite{song2019cvpr,chen2021iccv,kim2022cvpr} employ external memory to store domain-invariant class representations for image classification and semantic segmentation.

We propose the first work to solve DG in crowd counting. We build our work on top of MLDG~\cite{li2018aaai} where no target data/labels are required. Furthermore, we introduce new domain-invariant and -specific crowd memory modules. Memory vectors employed in~\cite{song2019cvpr,chen2021iccv,kim2022cvpr} normally correspond to explicit categorical concepts~(\eg object class, person id). They are initialized and iteratively updated by learned image features in the network. \cite{wu2021mm} have utilized this type of memory to build a  crowd counting model for DA where each memory vector is defined as the mean head feature per image. Their model does not address the varying size and number of heads in crowd images, hence memory vectors can be subject to noise. Unlike these works, we do not relate our memory vectors to explicit crowd concepts; instead, we initialize them as a set of random vectors. With properly devised loss functions, our memory modules are implicitly learned as sets of basis to re-encode image features into either domain-invariant or -specific features. 

%% file: sec3_method.tex
\input{Fig/overview}

\section{Method}


\subsection{Overview}
\label{sec:overview}
\noindent \textbf{Problem Formulation.} In this work, we are given one source dataset $\mathcal S$ consisting of crowd images with corresponding labels (head center annotations in each image). Our aim is to train a crowd counting model on $\mathcal S$ which can generalize well to any target (unseen) domain without the need of target data or further model updating. 

\noindent \textbf{Method Overview.} We illustrate our method in Fig.~\ref{fig:overview}. In the left, the base model has an encoder-decoder architecture where we extract and concatenate three-scale image features from it. Next, in order to separate between the domain-invariant (DI) and -specific (DS) information in the image feature, we introduce the domain-invariant crowd memory (DICM, Sec.~\ref{sec:agnostic}) and domain-specific crowd memory (DSCM, Sec.~\ref{sec:specific}) to re-encode the feature. The two modules are implemented in two separate branches, and each consists of the DI/DS unit followed by the DICM/DSCM module. In each branch, we first process the image feature through the DI/DS unit and then compute its similarities to memory vectors in DICM/DSCM. The re-encoding is thus realized as a weighted combination of memory vectors based on the similarity values. Notice DSCM has multiple sets of memory vectors, and each only re-encodes images belonging to a certain domain. We devise two types of loss functions, \ie feature reconstruction and orthogonal losses on the re-encoded domain-invariant and -specific features, specified in Sec.~\ref{sec:optim}. The domain-invariant feature is used for crowd density estimation where pixel-level density estimation loss is applied.  
The overall model optimization follows the meta-learning manner where we divide the source set $\mathcal S$ into multiple subsets to simulate the meta-train and -test sets in meta-learning. In Sec.~\ref{sec:dynamicdivision}, we introduce a dynamic sub-domain division scheme to cluster sub-domains and dynamically refine them during meta-learning. The meta-learning optimization is detailed in Sec.~\ref{sec:optim}.

\subsection{Domain-Invariant Crowd Memory}
\label{sec:agnostic}

The DI branch consists of the DI unit and the DICM module. DI unit has a 1$\times$1 convolution layer. It serves as a promoter to initially promote the domain-invariant information in the image feature. The transformed feature is later re-encoded via DICM to obtain the domain-invariant feature.  DICM is initialized as $M$ memory vectors $\{v_m\}$, each with the dimension of $C$. As discussed in Sec.~\ref{sec:II-dg}, each memory vector is normally associated with a concrete categorical concept depending on the task~\cite{song2019cvpr,chen2021iccv,kim2022cvpr}.  Nonetheless, we argue that it is rather difficult to explicitly determine concepts for memory vectors in crowd counting: density values are continuous, generating memory representation for each density value would be impossible; dividing the continuous density into discrete density levels may also suffer from quantization errors. To bypass this problem, our memory vectors are randomly initialized; and through properly defined feature re-encoding operation (below) and loss functions (Sec.~\ref{sec:optim}), they can automatically learn to store what is beneficial for domain-invariant density estimation. 

\textit{Feature re-encoding via DICM.} 
Given the input feature $f \in \mathbb{R}^{H\times W\times C}$ for DICM, we take every of its pixel-level vectors, \eg $f_{j} \in \mathbb{R}^{1\times C}$ at $j$-th pixel, to compute its similarities to memory vectors $\{v_m\}$ in DICM.
By applying Softmax over these similarity values, we obtain a set of weights to perform a weighted combination of memory vectors: 

\begin{equation}
\label{eq:read}
    \tilde{f_j} = \sum^M_m v_m \cdot \text{Softmax}(f_j v_m^\intercal)  
\end{equation}
$\tilde{f_j}$ is the domain-invariant re-encoding of $f_j$. We iterate every pixel vector in $f$ to obtain the domain-invariant feature $\tilde{f}$. $\tilde f$ is fed into the density estimator for density prediction.


With the proposed feature reconstruction and orthogonal loss (Sec.\ref{sec:optim}), DICM automatically updates its memory vectors through gradient back-propagation during model training. It gradually learns a set of basis to represent generic crowd information across domains.  

\subsection{Domain-Specific Crowd Memory}
\label{sec:specific}
DICM re-encodes an image feature into the domain-invariant feature $\tilde f$. 
However, having DICM alone is insufficient to achieve this goal. The reason is simple, we do not know what a domain-invariant feature looks like so we do not have a ground truth for the model to learn for it. Nevertheless, what we do know is that domain-invariant feature repels its domain-specific counterpart. Hence, if we can also re-encode the domain-specific feature from the original image feature, it can be served as a very useful opposite force to optimize against $\tilde f$.  


Similar to the DI branch, the DS branch has a DS unit (1$\times$1 conv) which is used to promote the domain-specific information in the image feature. The transformed feature, $z$, is then re-encoded via DSCM. Like DICM, memory vectors in DSCM are also randomly initialized; unlike DICM, DSCM has multiple sets of memory vectors, and each set is only responsible for re-encoding images that belong to certain sub-domain. The size of each memory set is $N$ and the dimension of each memory vector is $C$.     

\emph{Feature re-encoding via DSCM}. For each image, we have a sub-domain label $d$ for it (see Fig.~\ref{fig:overview} and Sec.~\ref{sec:dynamicdivision}). This label corresponds to the index of the memory set in DSCM. Following the same way to DICM, we use the selected memory set to re-encode the image feature into domain-specific feature $\tilde{z}$. During back-propagation, only the selected memory set will be updated. As a result, each DSCM memory set gradually learns a basis of vectors to represent specific crowd information  of certain domain.

\subsection{Dynamic Sub-domain Division}
\label{sec:dynamicdivision}

Given the source dataset $\mathcal S$, 
the intra-dataset variance is implicitly encoded into the image latent features extracted from the network. An intuitive way for domain division is hence through the clustering of image features. The clustering quality relies on the representativeness of image features.
Since image features get better representative alongside the update of network parameters, we propose to dynamically cluster and refine sub-domains.

Looking at our network, we have two candidate feature positions, \ie $z$ and $\tilde z$, that both contain sub-domain information. $\tilde z$ is however not suitable for the clustering because it has already been encoded using a tailored memory bank for a certain sub-domain.  
The re-encoded features using different sets of memory vectors are clearly separated (see Fig.~\ref{fig:cluster}). Despite the update of network parameters, the clustering result would not change. In contrast, $z$ is apparently more suitable: domain-specific information has been promoted in it yet all features regardless of domains are still extracted with the same extraction method. With network updating, as a result of the driving force from $\tilde z$, $z$ gets more prominent for sub-domain division.  
In practice, we conduct K-Means clustering upon image features $z$ before training (pre-trained backbone on ILSVRC) and after every epoch during training. 
To avoid the large shift between previous and current {sub-domain label assignments over} clusters, we permute current assignments of sub-domain labels across clusters to achieve major agreement to previous ones. We solve this as a
Maximum Bipartite Matching problem using the Kuhn-Munkres algorithm~\cite{munkres1957algorithms}.  
 


\label{sec:metamem}

\subsection {Network Training}

\label{sec:optim}

We introduce two losses for the optimization of DICM and DSCM: feature reconstruction loss and feature orthogonal loss. The former enforces the similarity between re-encoded features and their pre-encoded counterparts, which enhances the diversification of memory vectors within a memory bank and prevents their collapses. The latter enforces the dissimilarity between domain-invariant features and their domain-specific counterparts, so as to distinct the information learned in them. 


\noindent \textbf{DI/DS Feature Reconstruction Loss. }
$\tilde f$ and $\tilde z$ are the re-encoded versions of $f$ and $z$, respectively. On one hand, the domain-invariant (-specific) information should be enhanced in $\tilde f$ ($\tilde z$) compared to $f$ ($z$); on the other hand, the essential image information should be maintained between $\tilde f$ and $f$ ($\tilde z$ and $z$), so that different image features can still be discriminated. Take the example of  $\tilde f \in \mathbb{R}^{H\times W\times C}$ and $f \in \mathbb{R}^{H\times W\times C}$, to enforce their similarity, we follow \cite{wu2021iccv} to first flatten their features into the space of $\mathbb{R}^{HW\times C}$ and compute their correlation matrix $R\in \mathbb{R}^{HW\times HW}$.  
We can design a loss to maximize diagonal entries of $R$ so that the pixel-level features in $\tilde f$ and $f$ are highly correlated.  



\emph{Hard predicted region reinforcement.} We use $\tilde f$ for crowd density estimation. Every pixel-level vector in $\tilde f$ corresponds to a region in the density map. For regions with large prediction errors, we reinforce their learning in the loss function. Specifically, we first compute the pixel-wise MSE loss between the predicted density map and ground truth to obtain an error map. Pixel-wise error values are locally integrated into evenly partitioned regions, each region corresponding to the spatial position of one pixel in $\tilde f$. Top $S$ regions with the largest error values are regarded as hard predicted regions. We use their indices to find corresponding pixel-level vectors in $\tilde f$ and reinforce them in the reconstruction loss. 


Overall, we write the DI/DS feature reconstruction loss as  
\begin{equation}
    L_\text{rec} = \sum_{diag  = {[1, HW]}}(1+ \mathds{I}(diag))\times CE(R_{diag})
\end{equation}
where $CE$ is a cross entropy loss to maximize the diagonal element in $R$ (\ie $R_{diag}$). 
$\mathds{I}$ is an indicator function. For $\tilde f$, $\mathds I = \mathds{1}\{diag\in \text{HR}\}$ indicating whether the position $diag$ in $\tilde f$ belongs to the hard region indices $\text{HR}$. For $\tilde z$, $\mathds I = \mathds{1}\{diag\in \emptyset\}$ because $\tilde z$ is not used for density estimation. 

\noindent \textbf{DI-DS Feature Orthogonal Loss.} To distinguish domain-invariant and -specific features in the embedding space, we introduce a DI-DS feature orthogonal loss to minimize the squared Frobenius norm of the pixel-level feature similarity between $\tilde f$ and $\tilde z$: 

\begin{equation}
    {L_\text{orth} = \sum_{diag=[1,HW]}(\left\|{\tilde{O}_{diag}}\right\|_F^2 + \left\|{O_{diag}}\right\|_F^2)}
\end{equation}
where $\tilde{O}$ denotes the correlation matrix between $\tilde{f}$ and $\tilde{z}$ while $\tilde{O}_{diag}$ denotes the diagonal elements of $\tilde{O}$. In practice, we find it works better to also apply this loss between $f$ and $z$; hence, similarly, we add the term for ${O_{diag}}$. 

\noindent \textbf{Overall Optimization.} With the pixel-level MSE Loss $L_\text{den}$ between the predicted crowd density map and ground truth,  
we have the following overall optimization objective:
\begin{equation}
\label{eq:overall}
    L = L_\text{den} + \lambda_\text{rec} L_\text{rec} + \lambda_\text{orth} L_\text{orth}
\end{equation}
Our optimization follows the gradient-based meta-learning in~\cite{li2018aaai}.
We split the meta-train/test sets based on the sub-domains obtained in Sec.~\ref{sec:dynamicdivision}. 
One training iteration consists of two stages: in the first stage, the original model is trained on meta-train set to compute the loss in (\ref{eq:overall}) (denoted as $L^\text{mt}$) and update model parameters to obtain an intermediate model; in the second stage, the intermediate model is tested on the meta-test set to compute the meta-test loss $L^\text{me}$. $L^\text{mt}$ and $L^\text{me}$ are combined to update the original model later. 
The model parameter updating is:
\begin{equation}
\begin{aligned}
\theta=\theta-\gamma(\nabla_{\theta}L^\text{mt}(\theta)+\nabla_{\theta'}L^\text{me}(\theta'))    
\end{aligned}
\end{equation}
where $\theta$, $\theta'$ denote original and intermediate model parameters, respectively; and $\gamma$ the learning rate. 


\noindent \textbf{Inference.} We use only the DI branch for inference.






%% file: Fig/overview.tex
\begin{figure*}[t]
\begin{center}

\includegraphics[width=\textwidth]{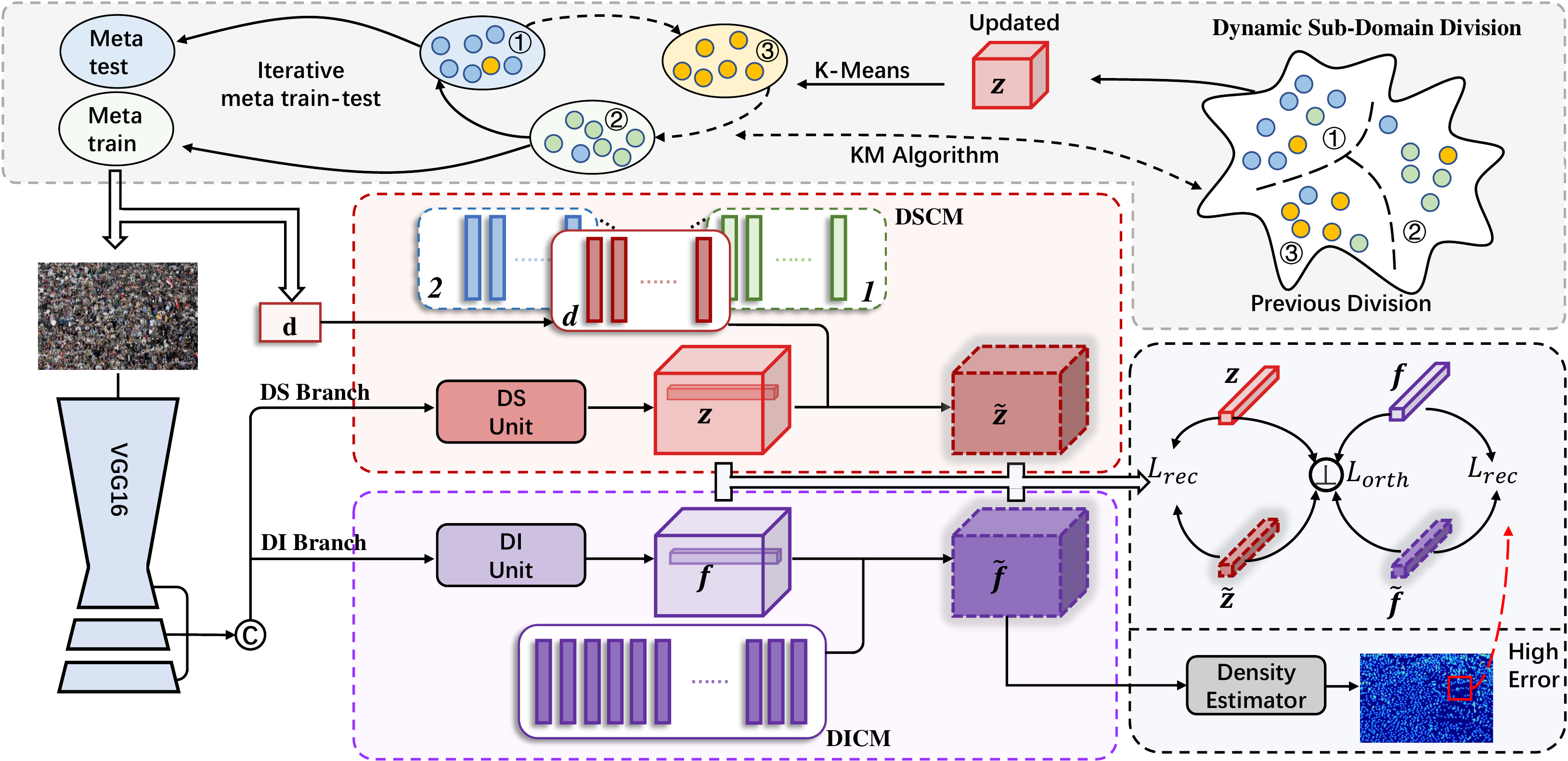}
\end{center}

\caption{{Flow chart of the proposed method.} Details of the data flow and DS/DI branches can be referred to Sec.~\ref{sec:overview}. }
%

\label{fig:overview}
\end{figure*}

%% file: sec4_experiments.tex
\section{Experiments}
\input{Tab/comparison}


\subsection{Experimental Setting}
\label{sec:expset}
\noindent \textbf{Datasets.} We evaluate our method on three crowd counting datasets: SHTech~\cite{zhang2016cvpr}, QNRF~\cite{idrees2018eccv}, NWPU~\cite{wang2020nwpu}.  SHTech contains two parts, SHA and SHB; the former is much denser than the latter on average. 
NWPU and QNRF are two large-scale datasets with diverse crowd scenarios, covering a wide range of crowd densities. 
We select one dataset as the source domain to train our model on its train set, and evaluate the trained model on the test sets of rest datasets.  Head center annotations are provided in every image to generate ground truth density map via Gaussian convolving with a fixed kernel size of 15. 






\noindent \textbf{Implementation Details.}
\label{sec:implement}
~We adopt the encoder-decoder architecture from~\cite{song2021aaai} as our base model. 
The density estimator is a convolution layer with 1$\times$1 filters. All convolution layers are followed by ReLU. We use Adam as optimizer with a fixed learning rate of $10^{-5}$. We augment training data by randomly cropping patches with a fixed size $320\times320$ from images and then apply random horizontal flipping. We train the model for 150 epochs, and each contains 100 iterations. In each iteration, we sample one image from each sub-domain and iterate the training by taking one as meta-test image and the rest as meta-train images.   
Empirically, we apply 3-means clustering to divide sub-domains in SHB and 4-means clustering in SHA and SHA+SHB. {For QNRF, we apply 8-means clustering.} 
The numbers of memory vectors $M$ and $N$ are set as 1024 and 256 for DICM and DSCM, respectively. Dimension $C$ is 256. In the hard region reinforcement (Sec.~\ref{sec:optim}), we select top $S = 2\% $ hard predicted regions over total regions per image.  Loss weights $\lambda_\text{rec}$ and $\lambda_\text{orth}$ in (\ref{eq:overall}) are set to 0.1 and 0.01, respectively.  
Parameters are tuned on the test set of the source domain. 

\noindent \textbf{Evaluation Metrics.}~We evaluate our method using two commonly used metrics in crowd counting, \ie mean absolute error (MAE) and mean squared error (MSE) between estimated and labeled crowd counts in each image. 



\subsection{Comparison with State of the Art}

In Table~\ref{tab:comparison}, we compare our result to state of the art by taking SHA or SHB as the source domain, SHB and QNRF or SHA and QNRF as target domains. We categorize comparisons in two groups: one without adaptation and one with it. The latter normally relies on the target data for model adaptation. Ours belong to the former.   

Comparable methods in the first group (without adaptation) are directly trained on the source domain and evaluated on the target domains. All of them are crowd counting methods except one, DG-MAN, which is a combination of a state of the art DG method~\cite{mansilla2021iccv} and  crowd counting method~\cite{lin2022cvpr}. ~\cite{mansilla2021iccv} proposes novel gradient agreement strategy Agr-Sum to mitigate gradient interference among multiple domains. We apply Agr-Sum to MAN~\cite{mansilla2021iccv} and denote the variant by DG-MAN. Our method outperforms state of the art in this group by a large margin. We point out that some methods may appear to be good in certain transfer setting yet are among the worst in other settings, indicating that they are not domain-general.

When comparing our method to state of the art in the second group (with adaptation), it is still on par with the best performance and is even superior in some entries (\eg SHA/SHB$\rightarrow$QNRF). This is very impressive because, unlike others, our method does not utilize target data for model parameter updating.

In Table~\ref{tab:comparison2}~(Left), we further conduct the experiment by training on SHA, SHB, SHA+SHB and testing on NWPU. We implement several very recent methods{~\cite{wang2020nips,song2021aaai,lin2022cvpr}} with their published codes or models. The comparison is without adaptation. NWPU is a very challenging large-scale dataset (1600 test images). Either by taking SHA or SHB as the source domain, our method shows best overall results. DMCount has its MSE on SHA$\rightarrow$NWPU slightly better than us, but the rest of its results is much worse than ours. Ours shows domain-general capability on every setting. 
Next, we increase the size of source domain by combining SHA and SHB. SHA and SHB are different in terms of their crowd distributions and environments, making the domain-general learning difficult. Nonetheless, our result is even better than training on the single SHA or SHB, and is significantly better than others trained on SHA+SHB. This shows the generalizability of our method worked on a large-scale mixed scenario dataset. 


Last, in Table~\ref{tab:comparison2}~(Right),  
we conduct an inverse generalization experiment as {QNRF}$\rightarrow$SHA/SHB. 
QNRF is larger and more challenging dataset than SHA/SHB as the source domain; ours on  it is still in general better than others, showing its strong generalizability.  


\input{Tab/comparison2}

\input{Tab/ablationCom}

\subsection{Ablation Study}
We conduct ablation study on SHA$\rightarrow$SHB / QNRF and SHB$\rightarrow$SHA / QNRF. 

\noindent \textbf{Baseline + MLDG.} Our baseline only consists of the base encoder-decoder and density estimator. Its result is in Table~\ref{tab:ablation}. We apply the meta-learning domain generalization (MLDG) to the baseline:  we use K-means to cluster the source domain into sub-domains based on image features extracted from the base encoder-decoder and simulate meta-learning among sub-domains.  
This brings large improvement on all experiments in Table~\ref{tab:ablation}, validating the necessitate of using DG technique for cross domain evaluation.  


\noindent \textbf{DI and DS Branches.} 
We first add the DI Branch with its feature reconstruction loss $L_\text{rec}$ to baseline + MLDG. The result, denoted by +DI-B by in Table~\ref{tab:ablation},  is consistently improved, which validates the effectiveness of using domain-invariant crowd memory (DICM) to re-encode image features. 
Next, we add the DS branch alongside the DI branch so both feature reconstruction loss $L_\text{rec}$ and orthogonal loss $L_\text{orth}$ are incorporated now. The result (+DS-B) shows that the performance is further significantly improved. This validates our argument that it is very important to incorporate the domain-specific crowd memory (DSCM) as the opposite force to optimize against DICM, so as to learn representative domain-invariant and -specific features. +DS-B is the full version of ours.  

\noindent \textbf{Feature Reconstruction and Orthogonal Losses.} 
We specifically investigate the effectiveness of two proposed losses. Table~\ref{tab:ablation} first shows the result without using feature reconstruction loss~(w/o $L_\text{rec}$). Severe degradation occurs on all settings, \eg +87.8 and +156.5 on MAE and MSE in SHB$\rightarrow$QNRF. 
Likewise, the performance is also degraded when we remove the feature orthogonal loss (w/o $L_\text{orth}$).


\input{Tab/ablationCluster}

\input{Fig/HRR}

\noindent \textit{Hard predicted region reinforcement.} We study the importance of the hard predicted region reinforcement in $L_\text{rec}$ for $\tilde f$ and $f$. Without using it (w/o HRR),  {all results get worse, }\eg an increase on MAE and MSE by 10.2 and 22.7 respectively on SHB$\rightarrow$QNRF.

\noindent \textit{Parameter variation for Hard Predicted Regions.} We vary the number of selected hard predicted regions ($S$) on SHA$\rightarrow$SHB. The results of MAE are presented in Fig.~\ref{fig:pramhrr}~(Left). Within a certain range, \eg $1\%\sim10\%$ of the total regions per image, the performance can be clearly improved, while $2\%$ (default) works the best.

\input{Tab/paramDSDD}

\input{Tab/paramMem}

\noindent \textbf{Dynamic Sub-domain Division (DSDD).} 
The DSDD scheme dynamically refines sub-domain division, its comparison is made to a static sub-domain division (SSDD): we use the original image feature from the base encoder-decoder to apply K-means clustering only once. Compared to DSDD, SSDD increases MAE and MSE in all settings in Table~\ref{tab:ablation}, which validates the effectiveness of DSDD.  


\noindent \textit{Parameter variation for DSDD.} We vary the number of clusters and frequency~(by epoch) of clustering, denoted by $K$ and $E$, in Table~\ref{tab:ablationcluster}. The best performance occurs at $K = 4$ and 3 for SHA and SHB, respectively. Meanwhile, re-clustering after each epoch is optimal, $E = 1$. 

\noindent \textit{Image features for DSDD.} Regarding the image features ($z$) used for dynamic sub-domain division (DSDD), 
one alternative way is to re-encode each image with different sets of memory vectors in DSCM and use the average feature to represent this image. We denote this by (DSDD w/ avg $\tilde z$) in Table~\ref{tab:ablationf}. It performs much worse than using $z$ for DSDD. In addition, we also choose features ($f$ and $\tilde f$) from the DI-branch. The results (DSDD w/ $f$ and w/ $\tilde f$), in Table~\ref{tab:ablationf} are much inferior to ours. $f$ and $\tilde f$ contain little sub-domain information, hence are not suitable for DSDD. 

\noindent \textit{Clustering methods for DSDD.} We replace K-Means with Gaussian Mixture Clustering(GM) and Spectral Clustering(SC). For each, we also vary the number of clusters $K$. We show the performance on SHA$\rightarrow$SHB / QNRF in Fig.~\ref{fig:pramhrr}~(Right). The best results all occur at $K = 4$ (our default). Using K-Means appears to be slightly better than using GM and SC.


\input{Fig/clustervis}

\noindent \textbf{Parameter Variation on DICM/DSCM.} We report the parameter variation for different number of memory vectors in DICM and DSCM, respectively. They are denoted by $(M, N)$ in  Table~\ref{tab:ablationMN}. 

First, we fix $N$ for DSCM and conduct experiments on different $M$ in DICM. Either increasing or decreasing $M$ from 1024 would cause the performance decrease.  $M=1024$ performs best. Next, with the fixed optimal value $M = 1024$, we vary $N$ for each memory set of DSCM. Among all the choices, $N=256$ appears the best. 

\input{Fig/visden}

\input{Fig/feature}

\noindent \textbf{t-SNE Visualization.} We visualize all the re-encoded domain-invariant and -specific features ($\tilde f$ and $\tilde z$) from SHB using t-SNE visualizations~\cite{van2008visualizing}. We denote them by $\tilde f_\text{DICM}$, $\tilde z_\text{DSCM1}$, $\tilde z_\text{DSCM2}$, $\tilde z_\text{DSCM3}$ in Fig.~\ref{fig:cluster} where the latter three are to signify domain-specifc features from three sub-domains, respectively. We can see that 1) $\tilde z_\text{DSCM1}$, $\tilde z_\text{DSCM2}$, $\tilde z_\text{DSCM3}$ are well separated as they are encoded with different sets of memory vectors; 2)  domain-invariant features $\tilde f_\text{DICM}$ from different sub-domains are clustered together and are away from domain-specific features $\tilde z_\text{DSCM1}$, $\tilde z_\text{DSCM2}$, $\tilde z_\text{DSCM3}$.  
We have also tried to draw the visualization without using $L_\text{rec}$ and $L_\text{orth}$ in our model: the separation among $\tilde f_\text{DICM}$, $\tilde z_\text{DSCM1}$, $\tilde z_\text{DSCM2}$, $\tilde z_\text{DSCM3}$ is no longer observed; and multiple samples of $\tilde z_\text{DSCM1}$, $\tilde z_\text{DSCM2}$ and $\tilde z_\text{DSCM3}$ are collapsed at some points.

\noindent \textbf{Feature Visualization.} Given an image, we also use channel-wise average pooling to collapse its re-encoded domain-specific and -invariant features ($\tilde z$ and $\tilde f$) into 2D feature maps ($H\times W$) for visualization. We observe in Fig.~\ref{fig:feature} that the former feature map focuses more on domain-specific style information such as background buildings and facilities; in contrast, the latter feature map focuses more on domain-invariant density information such as human heads. This validates the purpose of our design.


%

%% file: Tab/comparison.tex
\begin{table*}[!htb]
\centering

	\begin{tabular}{c|c|cc|cc|cc|cc}

		\toprule
		\multirow{2}{*}{Dataset}  & \textit{Source}  & \multicolumn{4}{c|}{SHA}  & \multicolumn{4}{c}{SHB}\\\cmidrule{2-10}  
		                    & \textit{Target} & \multicolumn{2}{c|}{ SHB} & \multicolumn{2}{c|}{ QNRF} & \multicolumn{2}{c|}{ SHA} & \multicolumn{2}{c}{ QNRF}\\
		
		\midrule
		Method & \textit{Adaptation} & MAE & MSE & MAE & MSE & MAE & MSE & MAE & MSE\\
	\midrule
	    MCNN~\cite{zhang2016cvpr} & \xmark & 85.2 & 142.3 & -- & -- & 221.4 & 357.8 & -- & --\\
	    DSSINet~\cite{liu2019iccv} & \xmark & 21.7 & 37.6 & 198.7 & 329.4 & 148.9 & 273.9 & 267.3 & 477.6\\
	    BL~\cite{ma2019iccv} & \xmark & {15.9} & {25.8} & 166.7 & 287.6 & 138.1 & 228.1 & 226.4 & 411.0 \\
	    DMCount~\cite{wang2020nips} &\xmark & 23.1 & 34.9& 134.4& 252.1& 143.9 & 239.6 & {203.0} & 386.1\\
	    D2CNet~\cite{cheng2021tip} & \xmark & 21.6 & 34.6 & {126.8} & {245.5} & 164.5 & 286.4 & 267.5 &486.0\\
	    SASNet~\cite{song2021aaai} & \xmark & 21.3& 33.2& 211.2& 418.6& {132.4}& {225.6}& 273.5& 481.3\\
	   MAN~\cite{lin2022cvpr} & \xmark & 22.1 & 32.8 & 138.8 & 266.3 & 133.6& 255.6 & {209.4} & {378.8}\\
	   DG-MAN~\cite{mansilla2021iccv} & \xmark & 17.3 & 28.7 & 129.1 & {238.2} & {130.7} & {225.1} & {182.4} & {325.8}\\
	   \textbf{Ours}  & \xmark &  \textbf{12.6} & \textbf{
	   24.6}  & \textbf{119.4} & \textbf{216.6} & \textbf{121.8} & \textbf{203.1} & \textbf{179.1} & \textbf{316.2} \\

	\midrule
	    Cycle GAN~\cite{zhu2017iccv} & \cmark &25.4 & 39.7 & 257.3 & 400.6 & 143.3 & 204.3 & 257.3 & 400.6 \\
	    SE CycleGAN~\cite{wang2019cvpr}& \cmark& 19.9 &28.3 & 230.4 & 384.5 & 123.4 & 193.4 & 230.4 & 384.5\\
	    SE+FD~\cite{han2020icassp}& \cmark & 16.9 & 24.7 & 221.2 & 390.2 & 129.3 & \underline{187.6} & 221.2 & 390.2\\
	    RBT~\cite{liu2020acmmm} & \cmark & 13.4 & 29.3 & 175.0 & 294.8 & \underline{112.2} & 218.2 & 211.3 & 381.9\\
	    C$^2$MoT~\cite{wu2021mm}&\cmark & \underline{12.4} & \underline{21.1} & \underline{125.7} & \underline{218.3} & 120.7 & 192.0 & \underline{198.9} & \underline{368.0}\\


		\bottomrule
	\end{tabular}
	\caption{Comparison with state of the art on SHA, SHB and QNRF. {All comparable results are copied from~\cite{wu2021mm}, except for DMCount, SASNet, MAN, and DG-MAN. We evaluate the latter four using their  published codes.} }
	\label{tab:comparison}	

\end{table*}

%% file: Tab/comparison2.tex
\begin{table*}[!t]

\centering

\begin{minipage}{0.68\textwidth}
	\begin{tabular}{c|cc|cc|cc}

		\toprule
        \textit{Source} & \multicolumn{2}{c|}{SHA} & \multicolumn{2}{c|}{SHB} & \multicolumn{2}{c}{SHA+SHB}\\
        \midrule
        \textit{Target} & \multicolumn{6}{c}{NWPU}\\
        \midrule
        Method & MAE & MSE & MAE & MSE & MAE & MSE \\
        \midrule
        DMCount~\cite{wang2020nips} & 146.9& \textbf{563.8}& 191.6& 747.4& 144.6& 592.8\\
        SASNet~\cite{song2021aaai} & 158.8 & 588.0 & 195.7 & 716.8 & 155.3 & 583.6 \\
        MAN~\cite{lin2022cvpr} & 148.2 & 586.5 & 193.6 & 802.5 & 147.8 & 605.3\\
        \midrule
        \textbf{Ours} & \textbf{143.1} & 567.6 & \textbf{175.0} & \textbf{688.6} & \textbf{139.6}& \textbf{553.6} \\

		\bottomrule
	\end{tabular}

\end{minipage}
\hfill
\begin{minipage}{0.3\textwidth}
    	\begin{tabular}{cc|cc}

		\toprule
         \multicolumn{4}{c}{QNRF}\\
        \midrule
        \multicolumn{2}{c|}{SHA} & \multicolumn{2}{c}{SHB} \\
        \midrule
        MAE & MSE & MAE & MSE\\
        \midrule
        73.4 & 135.1 & 14.3 & 27.5\\
        73.9 & 116.4 &13.0 &22.1\\
        \textbf{67.1} & 122.1 & 12.5 & 22.2\\
        \midrule
        67.4 & \textbf{112.8} & ~\textbf{12.1} & \textbf{20.9}\\

		\bottomrule
	\end{tabular}

\end{minipage}
	\caption{Left: Comparison with state of the art on NWPU. Right:  Inverse Generalization from QNRF .}
	\label{tab:comparison2}	
	
\end{table*}

%% file: Tab/ablationCom.tex
\begin{table*}[!ht]
    \centering
	\begin{tabular}{c|cc|cc|cc|cc}
    
		\toprule
		\textit{Source}  & \multicolumn{4}{c|}{SHA}  & \multicolumn{4}{c}{SHB}\\
		\midrule
		 \textit{Target} & \multicolumn{2}{c|}{ SHB} & \multicolumn{2}{c|}{ QNRF} & \multicolumn{2}{c|}{ SHA} & \multicolumn{2}{c}{ QNRF}\\
		
		\midrule
		Method & MAE & MSE & MAE & MSE & MAE & MSE & MAE & MSE\\
	\midrule
        Baseline & 22.0& 37.3& 240.2& 469.5 & 164.9 & 273.9 & 342.6 & 607.7\\
        +MLDG & 17.1 & 28.3 & 132.4 & 242.9 & 160.1 & 261.0 & 243.5 & 425.1\\
        +DI-B & 13.6 & 25.9 & 127.7 & 242.6 & 145.9 & 249.8& 233.9 & 394.2\\
        \textbf{+DS-B (Ours)}& \textbf{12.6}& {\textbf{24.6}}& \textbf{119.4} & \textbf{216.6} & \textbf{121.8} & \textbf{203.1}& \textbf{179.1} & \textbf{316.2} \\
        \midrule
		w/o $L_\text{rec}$ & 15.4 & 29.0 & 132.8& 253.0& 175.6 &299.6 & 266.9 & 472.7\\
		w/o $L_\text{orth}$ & 17.8& 31.4& 132.2 & 249.1 & 148.6 & 259.1 & 219.8 & 393.0\\
		w/o HRR & 13.4 & 25.3 & 124.0 & 233.5 & 127.8 & 215.4 & 189.3 & 338.9\\
		w/ SSDD &  14.0 & 25.9  & 125.6 & 237.7 & 128.3 & 219.5 & 198.8 & 345.4\\

		\bottomrule
	\end{tabular}
	\caption{Ablation study on our proposed components.}
	\label{tab:ablation}
\end{table*}

%% file: Tab/ablationCluster.tex
\begin{table}[!t]
    \centering
    \setlength{\tabcolsep}{3mm}
	\begin{tabular}{c|c|c|c|c}

		\toprule
		\textit{Source}  & \multicolumn{2}{c|}{SHA}  & \multicolumn{2}{c}{SHB}\\
		\midrule
		 \textit{Target} &  SHB &  QNRF & SHA & QNRF\\
		 \midrule
		 K=2 & 18.3& 141.4 & 127.5& 194.7\\
		 K=3 & 14.6 & 128.7& \textbf{121.8} & \textbf{179.1} \\
		 K=4 &\textbf{12.6} & \textbf{119.4}& 132.4 & 191.9\\
		 K=5 & 16.5 & 121.3 & 136.3 & 193.2\\
		 \midrule
		 E=1 & \textbf{12.6} &\textbf{119.4} & \textbf{121.8}&\textbf{179.1}\\
		 E=2 &14.2 &133.4 & 131.0& 195.0\\
		 E=3 & 16.3 & 133.5 & 138.9& 211.1\\
		 E=5 & 19.1& 144.3& 143.6& 219.8\\

		\bottomrule
	\end{tabular}
	\caption{Parameter variation for DSDD.}
	\label{tab:ablationcluster}	
\end{table}

%% file: Fig/HRR.tex
\begin{figure}[t]
\begin{center}

\begin{minipage}{0.495\linewidth}
\includegraphics[width=\linewidth]{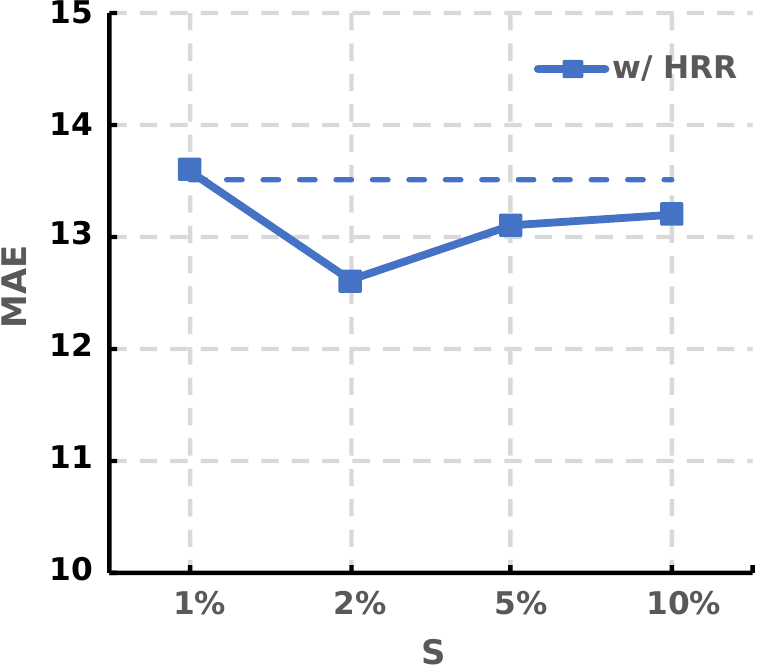}
\end{minipage}
\hfill
\begin{minipage}{0.495\linewidth}
\includegraphics[width=\linewidth]{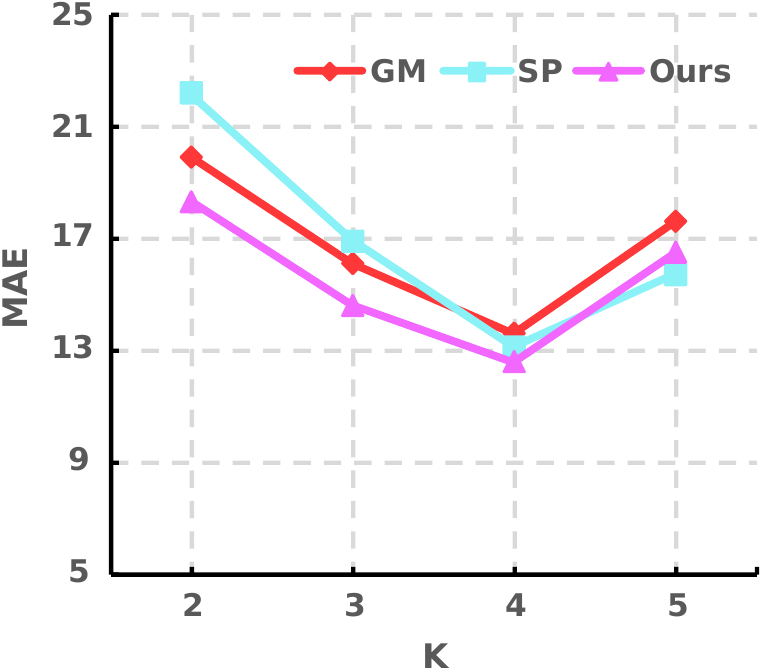}
\end{minipage}
\end{center}
\vspace{-5px}
\caption{Left: Parameter variation of hard predicted regions on SHA$\rightarrow$SHB. The dashed line indicates the performance without hard predicted region reinforcement (w/o HRR). Right: Different clustering methods for DSDD on SHA$
\rightarrow$SHB.  }
%

\label{fig:pramhrr}
\end{figure}

%% file: Tab/paramDSDD.tex
\begin{table*}[!htb]
    \centering
	\begin{tabular}{c|cc|cc|cc|cc}
    
		\toprule
		\textit{Source}  & \multicolumn{4}{c|}{SHA}  & \multicolumn{4}{c}{SHB}\\
		\midrule
		 \textit{Target} & \multicolumn{2}{c|}{ SHB} & \multicolumn{2}{c|}{ QNRF} & \multicolumn{2}{c|}{ SHA} & \multicolumn{2}{c}{ QNRF}\\
		
		\midrule
		Method & MAE & MSE & MAE & MSE & MAE & MSE & MAE & MSE\\
	\midrule
        \textbf{Ours}& \textbf{12.6}& \textbf{24.6}& \textbf{119.4} & \textbf{216.6} & \textbf{121.8} & \textbf{203.1}& \textbf{179.1} & \textbf{316.2} \\
        
        \midrule
		DSDD w/ $f$ & 15.4 & 29.2& 125.9& 240.0& 134.2& 225.5&201.0 & 365.8\\
		DSDD w/ $\tilde f$ &21.4 & 33.5 & 143.7& 274.9&143.7 & 247.8& 203.1 & 354.5\\
		DSDD w/ avg $\tilde z$& 16.5& 29.5& 135.8& 261.5& 127.0 & 217.2& 183.9 & 326.4\\

		\bottomrule
	\end{tabular}
	\caption{Ablation study for dynamic sub-domain division.}
	\label{tab:ablationf}
\end{table*}

%% file: Tab/paramMem.tex
\begin{table*}[!t]
    \centering
	\begin{tabular}{c|cc|cc|cc|cc}
    
		\toprule
		\textit{Source}  & \multicolumn{4}{c|}{SHA}  & \multicolumn{4}{c}{SHB}\\
		\midrule
		 \textit{Target} & \multicolumn{2}{c|}{ SHB} & \multicolumn{2}{c|}{ QNRF} & \multicolumn{2}{c|}{ SHA} & \multicolumn{2}{c}{ QNRF}\\
		
		\midrule
		(M,N) & MAE & MSE & MAE & MSE & MAE & MSE & MAE & MSE\\
	\midrule
        (1024, 256) &\textbf{12.6}& \textbf{24.6}& \textbf{119.4} & \textbf{216.6} & \textbf{121.8} & \textbf{203.1}& \textbf{179.1} & \textbf{316.2}\\

	\midrule
	    (256, 256) & 19.5 & 29.9 & 138.9 & 258.2 &130.5 & 228.3  & 221.4 & 387.6\\
	    (512, 256) & 16.0 & 29.5 & 137.1& 258.9& 124.7 & 209.3 & 185.9 & 333.9\\
	    (2048, 256) & 17.6& 31.2 &135.7 &260.5 &131.1 & 232.0 & 198.3 & 358.4\\
	    \midrule
        (1024, 128)& 18.1 & 30.7 & 141.0 & 237.0 & 128.0 & 222.3 & 203.1 & 370.2\\
        (1024, 512) &18.6 &31.2 &133.7 &245.0 & 146.2 & 253.9 & 223.0 & 409.1\\
        (1024, 1024) &21.1 &35.3 &135.0 &250.4 &141.0 & 237.0 & 202.9 & 365.0\\
        
    \bottomrule
	\end{tabular}
	\caption{Parameter variation ($M, N$) in DICM / DSCM.}
	\label{tab:ablationMN}
	\vspace{-10px}
\end{table*}

%% file: Fig/clustervis.tex
\begin{figure}[t]
\begin{center}

\begin{minipage}{0.495\linewidth}
\includegraphics[width=\linewidth]{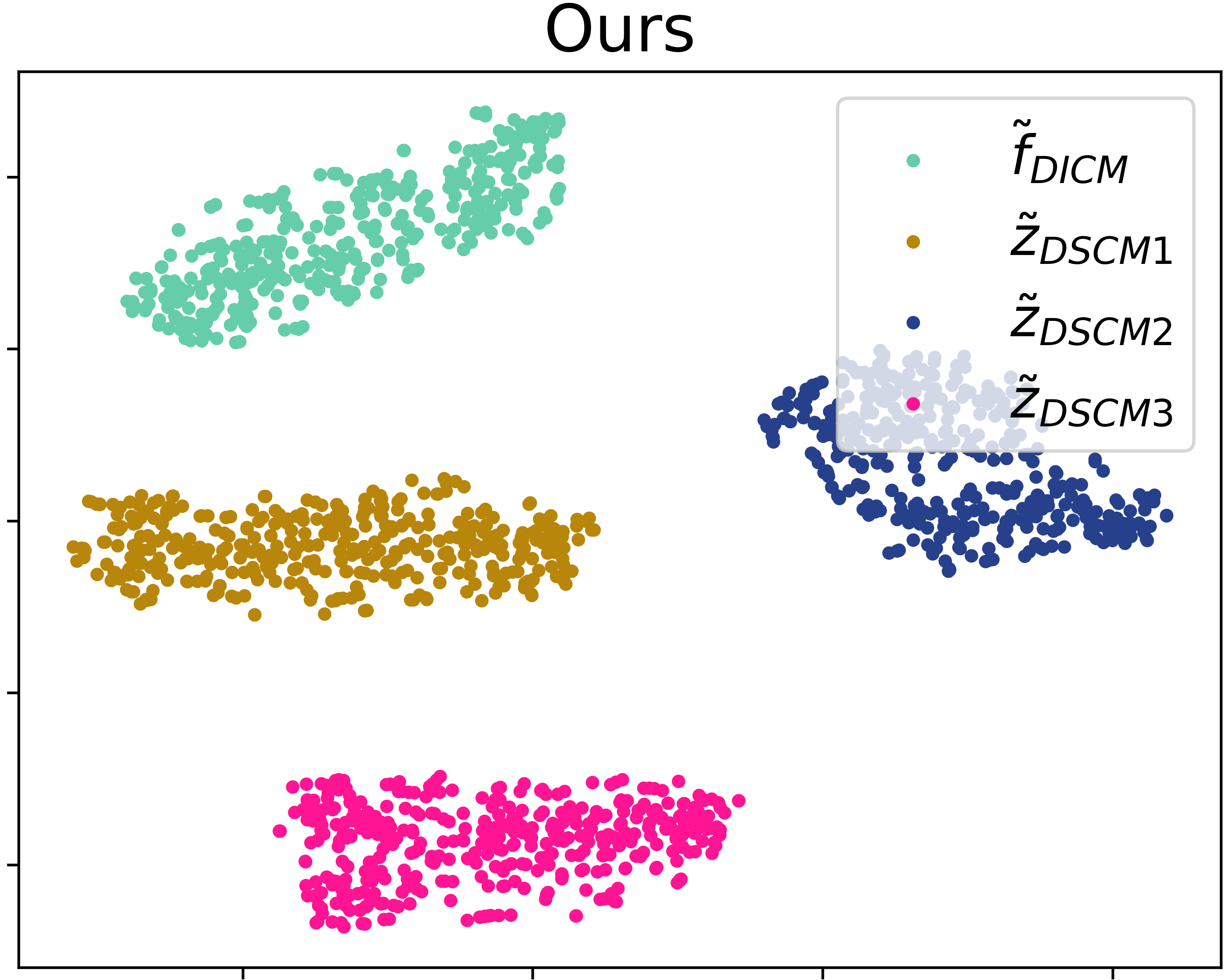}
\end{minipage}
\hfill
\begin{minipage}{0.495\linewidth}
\includegraphics[width=\linewidth]{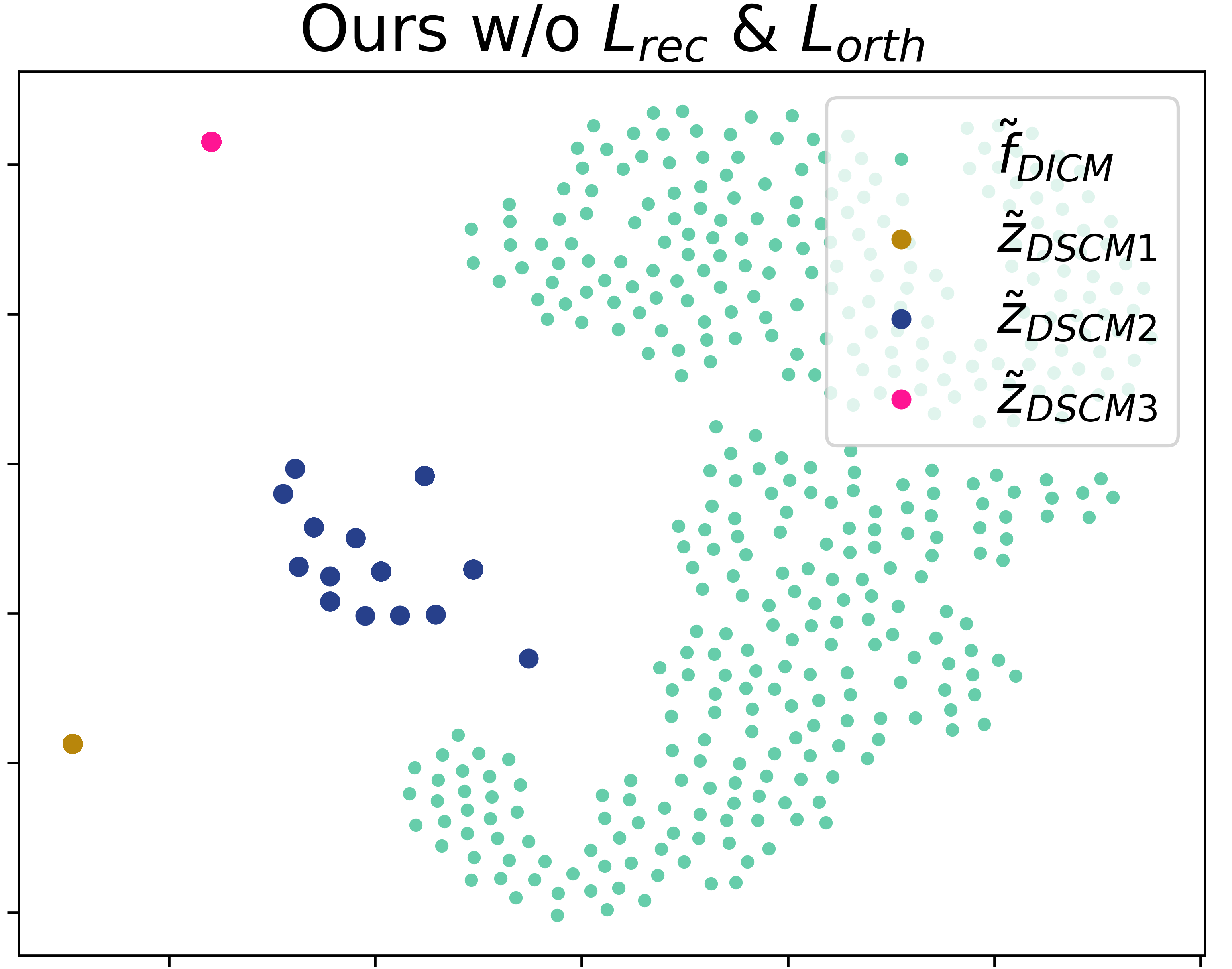}
\end{minipage}
\end{center}

\caption{t-SNE visualization of re-encoded features.}
%

\label{fig:cluster}
\end{figure}

%% file: Fig/visden.tex
\begin{figure}[t]
\begin{center}

\includegraphics[width=\linewidth]{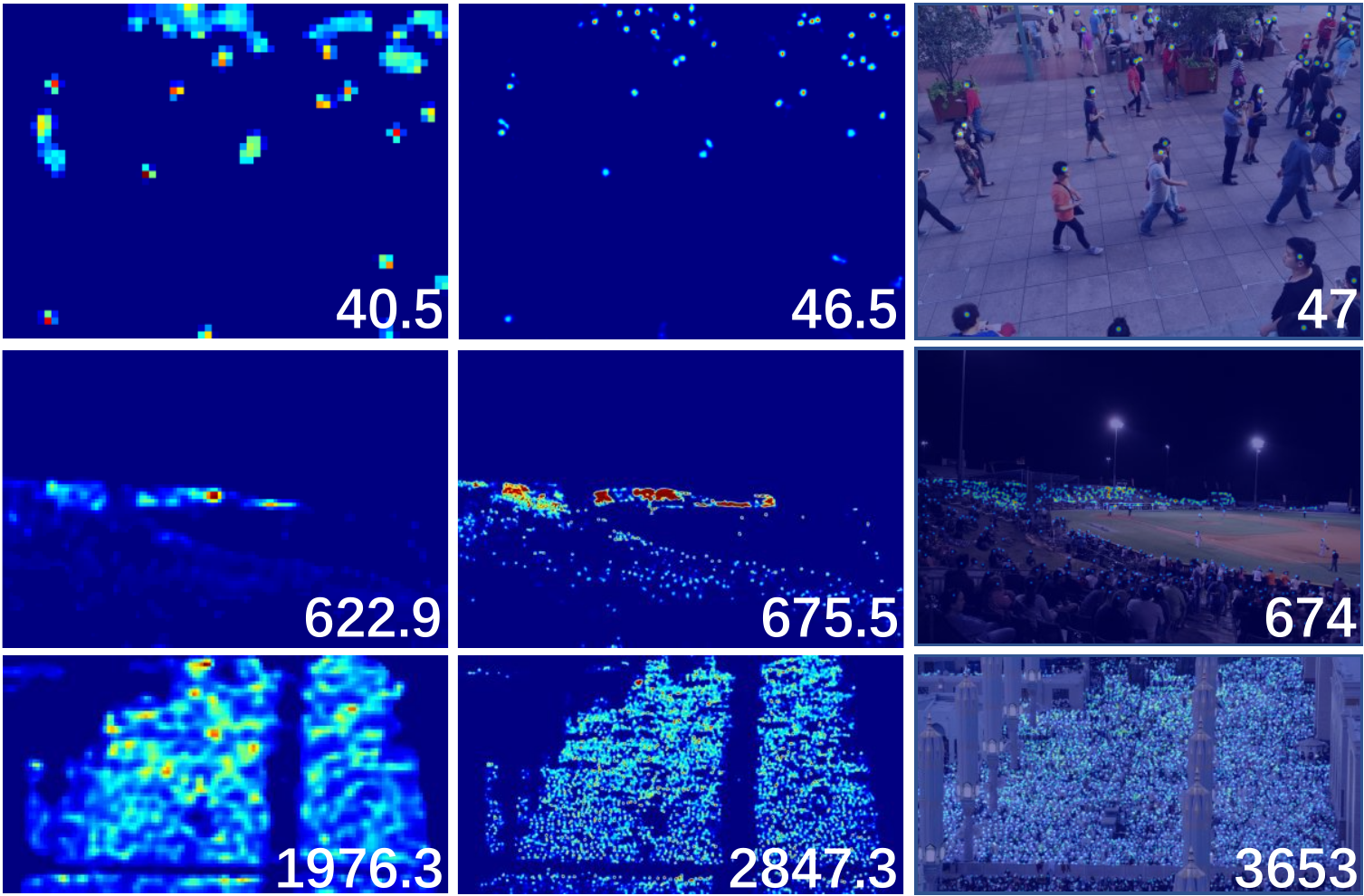}
\end{center}

\caption{Qualitative examples of MAN~\cite{lin2022cvpr}, ours, and ground truth~(from left to right).} 
%

\vspace{-8px}
\label{fig:visden}
\end{figure}

%% file: Fig/feature.tex
\begin{figure}[t]
\begin{center}

\includegraphics[width=\linewidth]{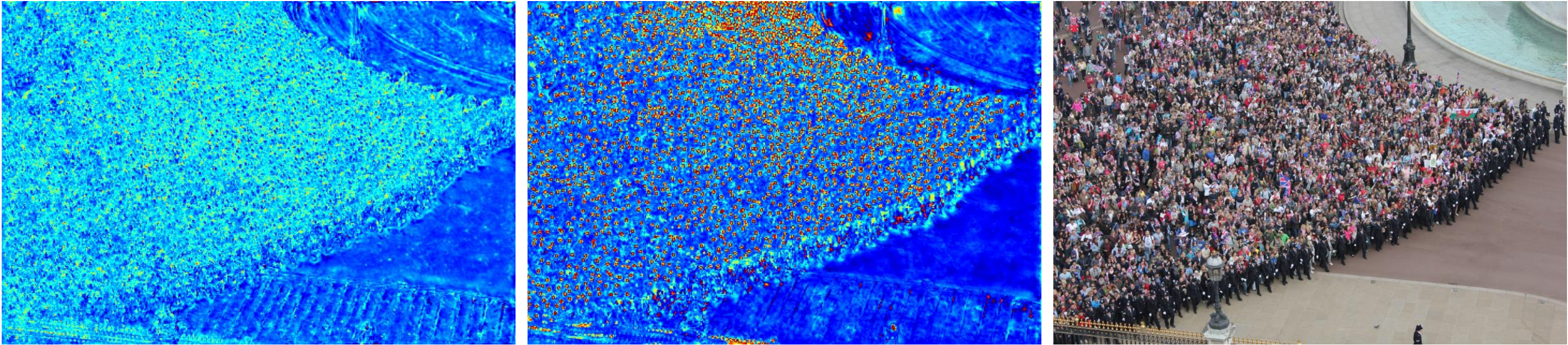}
\end{center}

\caption{Visualization of the domain-specific and -invariant feature maps of a given image~(from left to right).} 
%

\vspace{-8px}
\label{fig:feature}
\end{figure}

%% file: sec5_conclusion.tex
\section{Conclusion}

This paper studies the domain generalization in crowd counting for the first time. Our method is built on the meta-learning domain-generalization framework. We introduce domain-invariant and -specific crowd memory modules (DICM and DSCM) to extract disentangled domain-invariant/-specific features for each image. Feature reconstruction and orthogonal losses are introduced to distinguish the learning for DICM and DSCM. Finally, we design a dynamic sub-domain division strategy to dynamically refine sub-domain division for meta-learning. Experiments on several standard benchmarks validate the generalizability of our method. Future work will be improving the base model. 

%% file: sec6_ack.tex
\section*{Acknowledgments}
This work was partially supported by NSFC (No. 61828602) and EPSRC IAA grants. 

%% file: CameraReady.bbl
\begin{thebibliography}{55}
\providecommand{\natexlab}[1]{#1}

\bibitem[{Chen et~al.(2021{\natexlab{a}})Chen, Yan, Li, Li, Wang, Zuo, and
  Zhang}]{chen2021variational}
Chen, B.; Yan, Z.; Li, K.; Li, P.; Wang, B.; Zuo, W.; and Zhang, L.
  2021{\natexlab{a}}.
\newblock Variational attention: Propagating domain-specific knowledge for
  multi-domain learning in crowd counting.
\newblock In \emph{ICCV}.

\bibitem[{Chen et~al.(2021{\natexlab{b}})Chen, Wang, Pan, Yao, Tian, and
  Mei}]{chen2021iccv}
Chen, Y.; Wang, Y.; Pan, Y.; Yao, T.; Tian, X.; and Mei, T. 2021{\natexlab{b}}.
\newblock A style and semantic memory mechanism for domain generalization.
\newblock In \emph{ICCV}.

\bibitem[{Cheng et~al.(2021)Cheng, Xiong, Cao, and Lu}]{cheng2021tip}
Cheng, J.; Xiong, H.; Cao, Z.; and Lu, H. 2021.
\newblock Decoupled two-stage crowd counting and beyond.
\newblock \emph{TIP}.

\bibitem[{Dai et~al.(2021)Dai, Li, Liu, Tong, and Duan}]{dai2021cvpr}
Dai, Y.; Li, X.; Liu, J.; Tong, Z.; and Duan, L.-Y. 2021.
\newblock Generalizable person re-identification with relevance-aware mixture
  of experts.
\newblock In \emph{CVPR}.

\bibitem[{Du et~al.(2022)Du, Shi, Deng, and Zafeiriou}]{du2022redesigning}
Du, Z.; Shi, M.; Deng, J.; and Zafeiriou, S. 2022.
\newblock Redesigning Multi-Scale Neural Network for Crowd Counting.
\newblock \emph{arXiv preprint arXiv:2208.02894}.

\bibitem[{Gong et~al.(2022)Gong, Zhang, Yang, Dai, and Schiele}]{gong2022cvpr}
Gong, S.; Zhang, S.; Yang, J.; Dai, D.; and Schiele, B. 2022.
\newblock Bi-level Alignment for Cross-Domain Crowd Counting.
\newblock In \emph{CVPR}.

\bibitem[{Guo et~al.(2020)Guo, Zhu, Zhao, Cao, Lei, and Li}]{guo2020cvpr}
Guo, J.; Zhu, X.; Zhao, C.; Cao, D.; Lei, Z.; and Li, S.~Z. 2020.
\newblock Learning meta face recognition in unseen domains.
\newblock In \emph{CVPR}.

\bibitem[{Han et~al.(2020)Han, Gao, Yuan, and Wang}]{han2020icassp}
Han, T.; Gao, J.; Yuan, Y.; and Wang, Q. 2020.
\newblock Focus on semantic consistency for cross-domain crowd understanding.
\newblock In \emph{ICASSP}.

\bibitem[{He et~al.(2021)He, Ma, Wei, Hong, Ke, and Gong}]{he2021aaai}
He, Y.; Ma, Z.; Wei, X.; Hong, X.; Ke, W.; and Gong, Y. 2021.
\newblock Error-aware density isomorphism reconstruction for unsupervised
  cross-domain crowd counting.
\newblock In \emph{AAAI}.

\bibitem[{Hossain et~al.(2019)Hossain, Kumar, Hosseinzadeh, Chanda, and
  Wang}]{hossain2019bmvc}
Hossain, M.~A.; Kumar, M.; Hosseinzadeh, M.; Chanda, O.; and Wang, Y. 2019.
\newblock One-Shot Scene-Specific Crowd Counting.
\newblock In \emph{BMVC}.

\bibitem[{Idrees et~al.(2018)Idrees, Tayyab, Athrey, Zhang, Al-Maadeed,
  Rajpoot, and Shah}]{idrees2018eccv}
Idrees, H.; Tayyab, M.; Athrey, K.; Zhang, D.; Al-Maadeed, S.; Rajpoot, N.; and
  Shah, M. 2018.
\newblock Composition loss for counting, density map estimation and
  localization in dense crowds.
\newblock In \emph{ECCV}.

\bibitem[{Jiang et~al.(2020)Jiang, Zhang, Xu, Zhang, Lv, Zhou, Yang, and
  Pang}]{jiang2020cvpr}
Jiang, X.; Zhang, L.; Xu, M.; Zhang, T.; Lv, P.; Zhou, B.; Yang, X.; and Pang,
  Y. 2020.
\newblock Attention scaling for crowd counting.
\newblock In \emph{CVPR}.

\bibitem[{Kim et~al.(2022)Kim, Lee, Park, Min, and Sohn}]{kim2022cvpr}
Kim, J.; Lee, J.; Park, J.; Min, D.; and Sohn, K. 2022.
\newblock Pin the Memory: Learning to Generalize Semantic Segmentation.
\newblock In \emph{CVPR}.

\bibitem[{Li et~al.(2018)Li, Yang, Song, and Hospedales}]{li2018aaai}
Li, D.; Yang, Y.; Song, Y.-Z.; and Hospedales, T. 2018.
\newblock Learning to generalize: Meta-learning for domain generalization.
\newblock In \emph{AAAI}.

\bibitem[{Lin et~al.(2021)Lin, Yuan, Zhao, Sun, Wang, and Cai}]{lin2021iccv}
Lin, C.; Yuan, Z.; Zhao, S.; Sun, P.; Wang, C.; and Cai, J. 2021.
\newblock Domain-invariant disentangled network for generalizable object
  detection.
\newblock In \emph{ICCV}.

\bibitem[{Lin et~al.(2022)Lin, Ma, Ji, Wang, and Hong}]{lin2022cvpr}
Lin, H.; Ma, Z.; Ji, R.; Wang, Y.; and Hong, X. 2022.
\newblock Boosting Crowd Counting via Multifaceted Attention.
\newblock In \emph{CVPR}.

\bibitem[{Liu et~al.(2019{\natexlab{a}})Liu, Qiu, Li, Liu, Ouyang, and
  Lin}]{liu2019iccv}
Liu, L.; Qiu, Z.; Li, G.; Liu, S.; Ouyang, W.; and Lin, L. 2019{\natexlab{a}}.
\newblock Crowd counting with deep structured scale integration network.
\newblock In \emph{ICCV}.

\bibitem[{Liu, Durasov, and Fua(2022)}]{liu2022cvpr}
Liu, W.; Durasov, N.; and Fua, P. 2022.
\newblock Leveraging Self-Supervision for Cross-Domain Crowd Counting.
\newblock In \emph{CVPR}.

\bibitem[{Liu et~al.(2021)Liu, Li, Han, Zhang, Yang, Huang, and
  Sebe}]{liu2021iccv}
Liu, X.; Li, G.; Han, Z.; Zhang, W.; Yang, Y.; Huang, Q.; and Sebe, N. 2021.
\newblock Exploiting sample correlation for crowd counting with multi-expert
  network.
\newblock In \emph{ICCV}.

\bibitem[{Liu et~al.(2019{\natexlab{b}})Liu, Shi, Zhao, and
  Wang}]{liu2019point}
Liu, Y.; Shi, M.; Zhao, Q.; and Wang, X. 2019{\natexlab{b}}.
\newblock Point in, box out: Beyond counting persons in crowds.
\newblock In \emph{CVPR}.

\bibitem[{Liu et~al.(2020)Liu, Wang, Shi, Satoh, Zhao, and Yang}]{liu2020acmmm}
Liu, Y.; Wang, Z.; Shi, M.; Satoh, S.; Zhao, Q.; and Yang, H. 2020.
\newblock Towards unsupervised crowd counting via regression-detection
  bi-knowledge transfer.
\newblock In \emph{ACM-MM}.

\bibitem[{Liu et~al.(2022)Liu, Wang, Shi, Satoh, Zhao, and
  Yang}]{liu2022discovering}
Liu, Y.; Wang, Z.; Shi, M.; Satoh, S.; Zhao, Q.; and Yang, H. 2022.
\newblock Discovering regression-detection bi-knowledge transfer for
  unsupervised cross-domain crowd counting.
\newblock \emph{Neurocomputing}, 494: 418--431.

\bibitem[{Ma et~al.(2019)Ma, Wei, Hong, and Gong}]{ma2019iccv}
Ma, Z.; Wei, X.; Hong, X.; and Gong, Y. 2019.
\newblock Bayesian loss for crowd count estimation with point supervision.
\newblock In \emph{ICCV}.

\bibitem[{Mansilla et~al.(2021)Mansilla, Echeveste, Milone, and
  Ferrante}]{mansilla2021iccv}
Mansilla, L.; Echeveste, R.; Milone, D.~H.; and Ferrante, E. 2021.
\newblock Domain generalization via gradient surgery.
\newblock In \emph{ICCV}.

\bibitem[{Matsuura and Harada(2020)}]{matsuura2020aaai}
Matsuura, T.; and Harada, T. 2020.
\newblock Domain generalization using a mixture of multiple latent domains.
\newblock In \emph{AAAI}.

\bibitem[{Munkres(1957)}]{munkres1957algorithms}
Munkres, J. 1957.
\newblock Algorithms for the assignment and transportation problems.
\newblock \emph{Journal of the society for industrial and applied mathematics},
  5(1): 32--38.

\bibitem[{Ni et~al.(2022)Ni, Song, Luo, Zheng, Li, and Shen}]{ni2022cvpr}
Ni, H.; Song, J.; Luo, X.; Zheng, F.; Li, W.; and Shen, H.~T. 2022.
\newblock Meta Distribution Alignment for Generalizable Person
  Re-Identification.
\newblock In \emph{CVPR}.

\bibitem[{Reddy et~al.(2020)Reddy, Hossain, Rochan, and Wang}]{reddy2020wacv}
Reddy, M. K.~K.; Hossain, M.; Rochan, M.; and Wang, Y. 2020.
\newblock Few-shot scene adaptive crowd counting using meta-learning.
\newblock In \emph{WACV}.

\bibitem[{Reddy et~al.(2021)Reddy, Rochan, Lu, and Wang}]{reddy2021tmm}
Reddy, M. K.~K.; Rochan, M.; Lu, Y.; and Wang, Y. 2021.
\newblock AdaCrowd: unlabeled scene adaptation for crowd counting.
\newblock \emph{TMM}.

\bibitem[{Shi et~al.(2019)Shi, Yang, Xu, and Chen}]{shi2019revisiting}
Shi, M.; Yang, Z.; Xu, C.; and Chen, Q. 2019.
\newblock Revisiting perspective information for efficient crowd counting.
\newblock In \emph{CVPR}.

\bibitem[{Sindagi et~al.(2020)Sindagi, Yasarla, Babu, Babu, and
  Patel}]{sindagi2020eccv}
Sindagi, V.~A.; Yasarla, R.; Babu, D.~S.; Babu, R.~V.; and Patel, V.~M. 2020.
\newblock Learning to count in the crowd from limited labeled data.
\newblock In \emph{ECCV}.

\bibitem[{Song et~al.(2019)Song, Yang, Song, Xiang, and
  Hospedales}]{song2019cvpr}
Song, J.; Yang, Y.; Song, Y.-Z.; Xiang, T.; and Hospedales, T.~M. 2019.
\newblock Generalizable person re-identification by domain-invariant mapping
  network.
\newblock In \emph{CVPR}.

\bibitem[{Song et~al.(2021{\natexlab{a}})Song, Wang, Jiang, Wang, Tai, Wang,
  Li, Huang, and Wu}]{song2021iccv}
Song, Q.; Wang, C.; Jiang, Z.; Wang, Y.; Tai, Y.; Wang, C.; Li, J.; Huang, F.;
  and Wu, Y. 2021{\natexlab{a}}.
\newblock Rethinking counting and localization in crowds: A purely point-based
  framework.
\newblock In \emph{ICCV}.

\bibitem[{Song et~al.(2021{\natexlab{b}})Song, Wang, Wang, Tai, Wang, Li, Wu,
  and Ma}]{song2021aaai}
Song, Q.; Wang, C.; Wang, Y.; Tai, Y.; Wang, C.; Li, J.; Wu, J.; and Ma, J.
  2021{\natexlab{b}}.
\newblock To choose or to fuse? scale selection for crowd counting.
\newblock In \emph{AAAI}.

\bibitem[{Van~der Maaten and Hinton(2008)}]{van2008visualizing}
Van~der Maaten, L.; and Hinton, G. 2008.
\newblock Visualizing data using t-SNE.
\newblock \emph{JMLR}.

\bibitem[{Volpi et~al.(2018)Volpi, Namkoong, Sener, Duchi, Murino, and
  Savarese}]{volpi2018nips}
Volpi, R.; Namkoong, H.; Sener, O.; Duchi, J.~C.; Murino, V.; and Savarese, S.
  2018.
\newblock Generalizing to unseen domains via adversarial data augmentation.
\newblock \emph{NIPS}.

\bibitem[{Wan, Liu, and Chan(2021)}]{wan2021cvpr}
Wan, J.; Liu, Z.; and Chan, A.~B. 2021.
\newblock A generalized loss function for crowd counting and localization.
\newblock In \emph{CVPR}.

\bibitem[{Wang et~al.(2020{\natexlab{a}})Wang, Liu, Samaras, and
  Nguyen}]{wang2020nips}
Wang, B.; Liu, H.; Samaras, D.; and Nguyen, M.~H. 2020{\natexlab{a}}.
\newblock Distribution matching for crowd counting.
\newblock \emph{NIPS}.

\bibitem[{Wang et~al.(2020{\natexlab{b}})Wang, Gao, Lin, and Li}]{wang2020nwpu}
Wang, Q.; Gao, J.; Lin, W.; and Li, X. 2020{\natexlab{b}}.
\newblock NWPU-crowd: A large-scale benchmark for crowd counting and
  localization.
\newblock \emph{TPAMI}.

\bibitem[{Wang et~al.(2019)Wang, Gao, Lin, and Yuan}]{wang2019cvpr}
Wang, Q.; Gao, J.; Lin, W.; and Yuan, Y. 2019.
\newblock Learning from synthetic data for crowd counting in the wild.
\newblock In \emph{CVPR}.

\bibitem[{Wang et~al.(2021{\natexlab{a}})Wang, Han, Gao, and
  Yuan}]{wang2021tnnls}
Wang, Q.; Han, T.; Gao, J.; and Yuan, Y. 2021{\natexlab{a}}.
\newblock Neuron linear transformation: Modeling the domain shift for crowd
  counting.
\newblock \emph{TNNLS}.

\bibitem[{Wang et~al.(2021{\natexlab{b}})Wang, Luo, Qiu, Huang, and
  Baktashmotlagh}]{wang2021iccv}
Wang, Z.; Luo, Y.; Qiu, R.; Huang, Z.; and Baktashmotlagh, M.
  2021{\natexlab{b}}.
\newblock Learning to diversify for single domain generalization.
\newblock In \emph{ICCV}.

\bibitem[{Wu, Wan, and Chan(2021)}]{wu2021mm}
Wu, Q.; Wan, J.; and Chan, A.~B. 2021.
\newblock Dynamic Momentum Adaptation for Zero-Shot Cross-Domain Crowd
  Counting.
\newblock In \emph{ACM-MM}.

\bibitem[{Wu et~al.(2021)Wu, Shi, Lin, and Cai}]{wu2021iccv}
Wu, Z.; Shi, X.; Lin, G.; and Cai, J. 2021.
\newblock Learning meta-class memory for few-shot semantic segmentation.
\newblock In \emph{ICCV}.

\bibitem[{Xiong and Yao(2022)}]{xiong2022eccv}
Xiong, H.; and Yao, A. 2022.
\newblock Discrete-Constrained Regression for Local Counting Models.
\newblock \emph{arXiv:2207.09865}.

\bibitem[{Yan et~al.(2021)Yan, Li, Wang, Ren, and Zuo}]{yan2021tcsvt}
Yan, Z.; Li, P.; Wang, B.; Ren, D.; and Zuo, W. 2021.
\newblock Towards Learning Multi-domain Crowd Counting.
\newblock \emph{TCSVT}.

\bibitem[{Yang et~al.(2021)Yang, Cheng, Shiau, and Wang}]{yang2021nips}
Yang, F.-E.; Cheng, Y.-C.; Shiau, Z.-Y.; and Wang, Y.-C.~F. 2021.
\newblock Adversarial teacher-student representation learning for domain
  generalization.
\newblock \emph{NIPS}.

\bibitem[{Zhang et~al.(2022)Zhang, Yu, Yan, and Wang}]{zhang2022arxiv}
Zhang, H.; Yu, H.; Yan, Y.; and Wang, R. 2022.
\newblock Gated Domain-Invariant Feature Disentanglement for Domain
  Generalizable Object Detection.
\newblock \emph{arXiv:2203.11432}.

\bibitem[{Zhang, Shi, and Chen(2018)}]{zhang2018crowd}
Zhang, L.; Shi, M.; and Chen, Q. 2018.
\newblock Crowd counting via scale-adaptive convolutional neural network.
\newblock In \emph{WACV}.

\bibitem[{Zhang, Lin, and Chan(2021)}]{zhang2021cvpr}
Zhang, Q.; Lin, W.; and Chan, A.~B. 2021.
\newblock Cross-view cross-scene multi-view crowd counting.
\newblock In \emph{CVPR}.

\bibitem[{Zhang et~al.(2016)Zhang, Zhou, Chen, Gao, and Ma}]{zhang2016cvpr}
Zhang, Y.; Zhou, D.; Chen, S.; Gao, S.; and Ma, Y. 2016.
\newblock Single-image crowd counting via multi-column convolutional neural
  network.
\newblock In \emph{CVPR}.

\bibitem[{Zhao et~al.(2021)Zhao, Zhong, Yang, Luo, Lin, Li, and
  Sebe}]{zhao2021cvpr}
Zhao, Y.; Zhong, Z.; Yang, F.; Luo, Z.; Lin, Y.; Li, S.; and Sebe, N. 2021.
\newblock Learning to generalize unseen domains via memory-based multi-source
  meta-learning for person re-identification.
\newblock In \emph{CVPR}.

\bibitem[{Zhao et~al.(2020)Zhao, Shi, Zhao, and Li}]{zhao2020active}
Zhao, Z.; Shi, M.; Zhao, X.; and Li, L. 2020.
\newblock Active crowd counting with limited supervision.
\newblock In \emph{ECCV}.

\bibitem[{Zhou et~al.(2021)Zhou, Yang, Qiao, and Xiang}]{zhou2021arxiv}
Zhou, K.; Yang, Y.; Qiao, Y.; and Xiang, T. 2021.
\newblock Domain generalization with mixstyle.
\newblock \emph{arXiv:2104.02008}.

\bibitem[{Zhu et~al.(2017)Zhu, Park, Isola, and Efros}]{zhu2017iccv}
Zhu, J.-Y.; Park, T.; Isola, P.; and Efros, A.~A. 2017.
\newblock Unpaired image-to-image translation using cycle-consistent
  adversarial networks.
\newblock In \emph{ICCV}.

\end{thebibliography}
